\pdfoutput=1
\documentclass[11pt]{article}

\usepackage[final]{acl}

\usepackage{times}
\usepackage{latexsym}

\usepackage[T1]{fontenc}
\usepackage[utf8]{inputenc}

\usepackage{microtype}

\usepackage{inconsolata}

\usepackage{graphicx}

\usepackage{hyperref}       % hyperlinks
\usepackage{url}            % simple URL typesetting
\usepackage{booktabs}       % professional-quality tables
\usepackage{arydshln}
\usepackage{multirow}
\usepackage{float}
\usepackage{amsfonts}       % blackboard math symbols
\usepackage{amsmath}
\usepackage{amssymb}
\usepackage{yhmath}
\usepackage{bbm}
\usepackage{nicefrac}       % compact symbols for 1/2, etc.
\usepackage{microtype}      % microtypography
\usepackage{comment}
\usepackage{subcaption}
\usepackage[most]{tcolorbox}
\usepackage{soul}
\usepackage{enumitem}
\usepackage{amsthm}
\usepackage{thmtools}
\usepackage{csquotes}
\usepackage{arydshln}
\usepackage{algorithmicx}
\usepackage{algorithm}
\usepackage{algcompatible}

\AtBeginEnvironment{tcolorbox}{\small}

\usepackage{epigraph,varwidth}

\definecolor{recursion}{RGB}{61,133,198}
\definecolor{whileloop}{RGB}{106,168,79}
\definecolor{forloop}{RGB}{217,173,41}
\definecolor{White}{rgb}{1, 1, 1}
\definecolor{Periwinkle}{rgb}{0, 0, 0}
\definecolor{myblue}{rgb}{0.82, 0.94, 0.75}
\definecolor{mygreen}{rgb}{0.64, 0.76, 0.68}
\definecolor{myyellow}{rgb}{0.88, 0.54, 0.35}
\definecolor{mygreen}{rgb}{0.68, 0.9, 0.8}
\definecolor{mypink}{rgb}{0.2, 0.87, 0.2}
\definecolor{darkgreen}{RGB}{0,140,0}
\def\arrvline{\hfil\kern\arraycolsep\vline\kern-\arraycolsep\hfilneg}

\colorlet{LightGray}{White!98!Periwinkle}
\declaretheoremstyle[
    name=Protocol,
]{thmsty}

\newcommand{\score}{\textsc{NeoGauge}}
\newcommand{\framework}{\textsc{Denial Prompting}}
\newcommand{\dataset}{\textsc{NeoCoder}}
\title{
\vspace*{-0.5in}
{{\small \hfill NAACL'25}\\
\vspace*{.25in}} 
Benchmarking Language Model Creativity: \\A Case Study on Code Generation}

\author{%
  Yining Lu\thanks{Work done at the Johns Hopkins University.}\textcolor{blue}{$^{\iota}$} \quad Dixuan Wang\textcolor{blue}{$^{\alpha}$} \quad Tianjian Li\textcolor{blue}{$^{\alpha}$} \quad Dongwei Jiang\textcolor{blue}{$^{\alpha}$} \\
  {\bf \quad Sanjeev Khudanpur\textcolor{blue}{$^{\alpha}$} \quad Meng Jiang\textcolor{blue}{$^{\iota}$} \quad Daniel Khashabi\textcolor{blue}{$^{\alpha}$}}  \\
  \textcolor{blue}{$^{\iota}$}University of Notre Dame \quad \textcolor{blue}{$^{\alpha}$}Johns Hopkins University
}

\definecolor{darkred}{RGB}{200,0,0}
\definecolor{lightgreen}{RGB}{231,255,219}
\definecolor{lightred}{RGB}{252,231,234}
\definecolor{lightyellow}{RGB}{250,253,191}
\definecolor{DarkRed}{RGB}{130,25,0}
\definecolor{purplebg}{RGB}{229, 199, 244}

\begin{document}

\maketitle

\begin{abstract}
As LLMs become increasingly prevalent, it is interesting to consider how ``creative'' these models can be.
From cognitive science, creativity consists of at least two key characteristics: \emph{convergent} thinking (purposefulness to achieve a given goal) and \emph{divergent} thinking (adaptability to explore new environments or constraints) \citep{runco2003critical}. 
In this work, we introduce a framework for quantifying LLM creativity that incorporates the two design ingredients:  
(1) We introduce \framework{} which pushes LLMs to develop more creative solutions to a given problem by incrementally imposing new constraints on the previous solution, compelling LLMs to adopt new strategies. 
(2) We define  \score, a metric that quantifies both convergent and divergent thinking in the generated creative responses by LLMs.
We test the proposed framework on Codeforces problems, which serve as both a natural dataset for coding tasks and a collection of prior human solutions. We quantify \score{} for various proprietary and open-source models and find that even the most creative model, GPT-4, still falls short of demonstrating human-like creativity. 
We also experiment with advanced reasoning strategies (MCTS, self-correction, etc.) and observe no significant improvement in creativity. 
As a by-product of our analysis, we release \dataset{} dataset for reproducing our results on future models.\footnote{
% Our code and dataset are available at 
Our code and data: 
\href{https://github.com/JHU-CLSP/NeoCoder}{\tt github.com/JHU-CLSP/NeoCoder}
}
% \href{https://anonymous.4open.science/r/NeoCoder-4B0E/README.md}{https://anonymous.4open.science/r/NeoCoder-4B0E/README.md}
% }
\end{abstract}

\begin{comment}
\epigraph{
\textit{Creative achievement depended not only on the number of alternatives but also the generation of a single high-quality alternative.}
}{Mark Runco, Critical Creative Process~\cite{runco2003critical}}
\vspace{-2mm}
\end{comment}

\section{Introduction}
\label{sec: introduction}

Most recent works on LLM creativity evaluation focus on open-ended generation tasks, such as story-writing \citep{atmakuru2024cs4measuringcreativitylarge, gómezrodríguez2023confederacy, 10.1145/3613904.3642731, chakrabarty2024creativity}, paper abstract generation \citep{lu2024aihumanityssalieriquantifying}, and role-play discussion \citep{lu2024llmdiscussionenhancingcreativity}.
% LLMs have shown increasing performance across a wide variety of \yining{problem-solving} tasks \citep{cobbe2021gsm8k, huang2019cosmos, lu2023gear,  onoe2021creak, patel-etal-2021-nlp, rajpurkar2016squad, roemmele2011choice, ye2024analobench}.
However, the degree to which LLMs possess and utilize \emph{creativity} for problem-solving remains unclear. An automatic method for evaluating LLMs creativity could help developers better understand the emergence of model behaviors and serve as a design objective in solving complex real-world problems. 

However, despite the importance of creativity evaluation in problem-solving, only a few works have touched upon it \citep{delorenzo2024creativeval, tian2024macgyver} because of two major challenges: (1) eliciting diverse and creative solutions is difficult \citep{bronnec2024exploring, xu2024wizardlm, zhang2024improving}, and (2) there are no reliable and comprehensive quantitative measurements of LLM creativity. Below, we explain how we tackle these two challenges for evaluating LLM creativity in problem-solving settings.

\begin{figure}[t]
\centering
\includegraphics[width=0.45\textwidth]{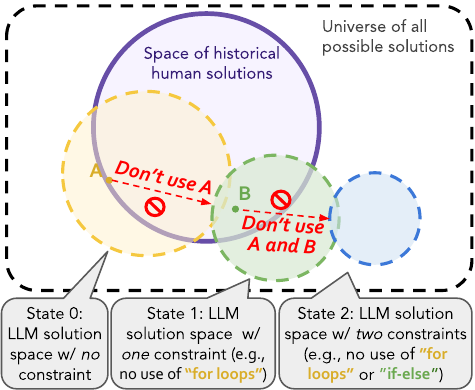}
\caption{An overview of how \framework{} encourages creative solutions. A solution space is a collection of all possible solutions at a certain state. \textcolor{forloop}{A}, \textcolor{whileloop}{B} indicate atomic techniques (e.g., for-loops, if-else, etc.) used in the solution.}
\label{fig: teaser figure}
\end{figure}

LLM generations are often repetitive and regurgitating training data \citep{holtzman2019curious, kirk2024understanding, tevet-berant-2021-evaluating, xu2024wizardlm, zhang2024forcing}, making it hard to elicit creative generations.
However, we argue that an effective creativity evaluation method should be based on the spectrum of maximal \textit{\textbf{creative solutions}} attained from LLMs. Therefore, we introduce \framework{} (\S\ref{subsec: dp framework}), a prompting method that iteratively ``denies'' one of the basic tools, techniques, or strategies used in the previous solution (e.g., \textcolor{forloop}{A: for loops} and \textcolor{whileloop}{B: if-else} in \autoref{fig: teaser figure}), thereby pushing LLM to think out-of-the-box and elicit creative generations to its fullest extent. 

Another challenge in creativity evaluation is to build a reliable and comprehensive quantitative measurement. We propose that such evaluation should be \textit{\textbf{state-aware}---adaptive to different contexts}, and \textit{\textbf{human-grounded}---comparing LLM-generated solutions to historical human solutions}. According to many cognitive studies, human creativity is viewed as taking place in the interaction with a person, environment, or another model~\citep{amabile1996creativity, csikszentmihalyi1996creativity, Csikszentmihalyi_1998, Feldman_1998, feldman1994changing, holyoak2005cambridge}.
Similarly, the essence of LLM creativity should also be captured from its interaction with the current state (state-aware) and past human knowledge background (human-grounded).
This understanding reveals that creativity evaluation should be dynamic, with an individual's creativity varying under different contexts.
For example, in \autoref{fig: teaser figure}, a solution at state $t=0$ probably will not be judged at the same creative level as one at state $t=2$, even if they solve the same problem. Because the latter solution is more likely to use novel techniques that humans hardly thought of, such as \textcolor{recursion}{C: Recursion}, to adapt to increasingly challenging constraints.

To address the second challenge, we propose \score{} score (\S\ref{subsec: creativity score})
which involves (1) verifying the correctness of the LLM-generated solution and whether it adheres to the specified constraints from \framework{} (convergent thinking), and (2) assessing solution novelty by contrasting it with techniques previously used in human solutions (divergent thinking). This aligns well with the arguments made by \citet{runco2003critical} that creative achievement depends on both the number of alternative solutions and the generation of high-quality alternatives. By considering both convergent \citep{lubart2001models, Sternberg1981IntelligenceAN, STERNBERG1982451, STERNBERG1989187} and divergent \citep{guilford1950creativity, holyoak2005cambridge, torrance1966torrance} creative thinking, \score{} not only offers a state-aware evaluation but grounds the evaluation in collective human knowledge through comparing the generated solutions with historical human solutions. 

In our experiments, we apply \framework{} on Codeforces,\footnote{\href{https://codeforces.com/problemset}{https://codeforces.com/problemset}} a challenging Text-to-Code task where model solutions can be automatically verified and allows comparison to substantial historical human solutions.\footnote{We provide detailed justifications for task choice in \S\ref{appendix: why codeforces}.} Specifically, we retrieve 199 latest problems from Codeforces along with 30 human solutions per problem that have successfully passed unit tests. We then run these problems on \framework{} to obtain our dataset \dataset{} which consists of original questions with sequences of temporally relevant and increasingly difficult constraints.
Examples of \dataset{} are provided in \autoref{table: examples of dataset}. We benchmark a broad range of LLMs on \dataset{} and calculate their \score{} scores. Additionally, we evaluate four reasoning strategies, MCTS \citep{zhang2023planning}, self-correction \citep{shinn2023reflexion}, planning \citep{jiang2023selfplanning}, and sampling \citep{chen2021evaluating}, on our dataset to study the correlation between augmented machine intelligence and creativity. 
In summary, our contributions are twofold: 
\begin{itemize}[leftmargin=*]
    \item We introduce \framework{} to elicit creative generations from LLMs and \score{} metric to evaluate LLM creativity in problem-solving that follows the two proposed protocols.
    \item We release a creativity benchmark \dataset{} and provide a thorough analysis of creativity on SOTA language models and reasoning strategies.
\end{itemize} 

\section{Background and Related Works}
We discuss the existing works on machine creativity evaluation. Then, we explain the concepts of divergent and convergent creativity in cognitive science which our evaluation incorporates.

\paragraph{Machine Creativity Evaluation.}
% \subsection{Machine Creativity Evaluation}
\label{subsec: machine creativity evaluation}
While the extensive studies on human creativity from psychological and cognitive science \citep{amabile1982social, finke1996creative, guilford1950creativity, mumford1991process, runco2003critical, sternberg1991investment, torrance1966torrance}, LLM creativity has received little attention. Existing works studying LLM creativity in problem-solving settings \citep{delorenzo2024creativeval, tian2024macgyver, zhu2024dyval}, however, tend to overlook two challenges: (1) eliciting creative LLM solutions, and (2) ensuring evaluation metrics are grounded and comprehensive. 

\citet{tian2024macgyver} have released a challenging real-world problem dataset to push LLM to think out-of-the-box, but they do not provide an automatic creativity evaluation method built upon their dataset. Additionally, their problems are constructed from a single constraint. In contrast, our \framework{} is formulated for multiple iterations of constraint detection and problem refinement, making the generations more creative and providing more states for creativity evaluation. 
Another concurrent work \citep{atmakuru2024cs4measuringcreativitylarge} also employs multiple constraints to facilitate creative generation; however, their evaluation primarily targets linguistic creativity \citep{lu2024aihumanityssalieriquantifying} and it is tested on open-ended story writing task.
\citet{zhu2024dyval} and \citet{xu2024wizardlm} design protocols to dynamically generate challenging problems with controllable constraints. However, their evaluation mainly focuses on accuracy rather than creativity. 

\citet{10.1145/3613904.3642731}, \citet{delorenzo2024creativeval}, and \citet{zhao2024assessing} introduce automatic evaluation pipelines to quantify the four subcomponents of creativity proposed in the Torrance Tests of Creative Thinking \citep{torrance1966torrance}: fluency, flexibility, originality, and elaboration. However, the test is originally designed to study human divergent creative thinking (\S\ref{subsec: divergent creative thinking}) and is unclear whether it applies to machine creativity. 
% Additionally, their evaluation methods hardly meet the above two proposed protocols. 

% \subsection{Divergent Creative Thinking}
\paragraph{Divergent Creative Thinking.}
\label{subsec: divergent creative thinking}
Divergent thinking is a cognitive process that involves exploring a multitude of potential applications for a given set of tools \citep{holyoak2005cambridge}. It typically occurs \textit{spontaneously} and \textit{randomly}, leading to numerous possible solutions. Extensive research \citep{amabile1982social, guilford1950creativity} has been conducted to study divergent creativity, including popular psychometric approaches such as the Unusual Uses Test \citep{guilford1950creativity}. These are designed to let examinees think of as many uses for a (common or unusual) object as possible. The underlying idea of stimulating creative solutions from constrained and unusual settings is also adopted in our \framework.

Divergent thinking can also be viewed through the lens of $\mathcal{P}$-creativity (\underline{P}sychological) and $\mathcal{H}$-creativity (\underline{H}istorical) defined by \citet{boden1994dimensions}. A valuable idea is $\mathcal{P}$-creative if the person in whose mind it arises could not have come up with it before. Furthermore, a valuable idea is $\mathcal{H}$-creative if it is $\mathcal{P}$-creative, and no one else in human history has ever had it before. $\mathcal{P}$-creativity measurement is embedded in the structure of \framework, where at each state, the LLM is prompted to come up with a brand new solution that it has never thought of before by imposing a new constraint. 
Therefore, we mainly consider $\mathcal{H}$-creativity measurement in our \score{} score, where we compare the model-generated solution with a set of collected human solutions to examine if it has ever been proposed in human history (i.e., the ratio of the region out of human solution space in \autoref{fig: teaser figure}). This makes our \score{} human-grounded and reflects the novelty from history.

% \subsection{Convergent Creative Thinking}
\paragraph{Convergent Creative Thinking.}
\label{subsec: convergent creative thinking}
Since the twenty-first century, more researchers have begun to accept the proposition that creative thought involves not merely the generation of many alternative solutions (divergent thinking) but also the identification of new \textit{feasible} solutions \citep{baer1994divergent, runco2003critical}. They frame this problem-solving process as convergent creative thinking and begin to examine how understanding human cognition and convergent thinking might be used to account for creative thought \citep{finke1996creative, mumford1991process, sternberg1991investment}. Several famous cognitive approaches that study the mental representation and process underlying convergent creative thinking \citep{lubart2001models} involve asking examinees to predict future states from past states using incomplete information \citep{Sternberg1981IntelligenceAN, STERNBERG1982451}, or solving the problems as though the counterfactual premises are true \citep{STERNBERG1989187, sternberg1989if}. All these tests share certain characteristics, such as always having a single best answer and asking examinees to think in unconventional ways.
In our work, besides computing $\mathcal{H}$-creativity for evaluating divergent thinking, our work also measures convergent creativity by verifying the feasibility of the generated solution: whether they are correct and following the given constraints. Our \score{} metric delivers a more comprehensive evaluation of machine creativity. 

\section{Constructing the \dataset{} Dataset}
We present \framework{} to stimulate creative responses from LLMs. 

\subsection{\framework{}: Eliciting Creative Generations from LLMs}
\label{subsec: dp framework}

Our purpose is to construct a pipeline that iteratively imposes constraints on previous solutions (e.g., disallowing the use of hashmaps) to force more creative solutions. The setup is as follows: given an input problem, we use a highly capable \textit{augmentation model} $\mathbf{P}_{\text{LM}}$ (e.g. {\tt GPT-4}) to generate solutions and scrutinize ``technique(s)'' used in the generated solution, then update the problem by imposing the detected technique as a constraint. We repeat this process $t$ times to obtain consecutive $t$ problems with increasingly hard constraints (\autoref{fig: dp} shows an example with $t=2$). 

Specifically, as shown in Algorithm \ref{alg: dp}, given a reasoning problem $x$ and an initial empty constraint list $\mathcal{C}_0 = \{\}$, 
we first let the augmentation model $\mathbf{P}_{\text{LM}}$ to generate an initial solution $y_1 \sim \mathbf{P}_{\text{LM}}(x)$ via a default problem-solving prompt and conversation history. We then use the same augmentation model $\mathbf{P}_{\text{LM}}$ to detect \textit{atomic techniques} (e.g., recursion, for loop, hashmaps, etc.), $\mathcal{T}_1=\{\tau^1, \tau^2, \cdots, \tau^i \}$, used in $y_1$ to solve $x$ with a technique detection prompt. Then, one technique is randomly sampled $\tau_1 \sim \mathcal{T}_1 \setminus \mathcal{C}_0$ to ensure it has never been used before as a constraint. Finally, we update the problem $x$ to $x\oplus \tau_1$ which explicitly  \textbf{prohibits} the use of the technique $\tau_1$ and update constraint list $\mathcal{C}_0$ to $\mathcal{C}_1 = \{\tau_1\}$.\footnote{We use $\oplus$ to indicate text concatenation.} This is the first iteration of \framework{}. We repeat the process to progressively obtain the overall constraint list $C_t = \{\tau_1, \tau_2,\cdots,\tau_t\}$. The prompts for \framework{} (including technique detection; used across all experiments) are in \autoref{appendix: prompts for denial prompting}. 

\begin{algorithm}[ht]
\small
      \caption{\framework{}}
        \begin{flushleft}
        \textbf{Input:} Input problem $x$, augmentation model $\mathcal{P}_{\text{LM}}$, max iterations $T$ \\
        \textbf{Output:} Constraint list $\mathcal{C}_T$
        \end{flushleft}
\begin{algorithmic}[1]
    \FOR {$t=1$ \textbf{to} $T$} 
    \STATEx \;\;\; \textcolor{darkgreen}{\# Response generation}
    \STATE $y_t \sim \mathbf{P}_{\text{LM}}(x\oplus\tau_1\oplus\cdots\oplus\tau_{t-1})$
    \vspace{0.1cm}
    \STATEx \;\;\; \textcolor{darkgreen}{\# Technique detection}
    \STATE $\mathcal{T}_{t} \sim \mathbf{P}_{\text{LM}}(y_t)$ 
    \vspace{0.1cm}
    \STATE $\tau_t \sim \mathcal{T}_t \setminus\mathcal{C}_{t-1}$ 
    \STATE $\mathcal{C}_{t} = \{\tau_1, \tau_2, \cdots, \tau_t\}$
    \ENDFOR
\end{algorithmic}
      \label{alg: dp}
\end{algorithm}

During \framework{}, we use a single conversation thread of $\mathbf{P}_{\text{LM}}$ to infer $y_t$ 
% Namely, we put $y_t \sim \mathbf{P}_{\text{LM}}(x\oplus\tau_1\oplus\cdots\oplus\tau_{t-1})$ (line 2 in Algorithm \ref{alg: dp}) over $t$ in one single thread, 
such that the model can utilize the trace of previous interactions (including problem statements, constraints, and LLM solutions from each iteration). 
In practice, we observe adding prior interactions in the context improves model generations. 
Conversely, when detecting solution techniques $\mathcal{T}_{t} \sim \mathbf{P}_{\text{LM}}(y_t)$ (line 4 in Algorithm \ref{alg: dp}), we disregard the context from previous conversation rounds to focus the responses solely on the most recent round.

\subsection{\dataset{} Dataset to Support Benchmarking LLM Creativity}
\label{appendix: results of dp}

\paragraph{Challenging problems.} To construct our creativity benchmark, we compile $n=199$ latest Codeforces problems. We chose problems with a difficulty of 800 (easiest level) since, in our preliminary experiments, we observed near-random performance on more challenging problems when using well-known open-source models.
Furthermore, we selected the recent data to prevent any memorization during pre-training~\cite{huang2023competition}.

\paragraph{Human solutions.}
For each problem, we extract $m=30$ correct human solutions per problem (total of $5.9$K human solutions).\footnote{We consider 30 human-annotated solutions to construct a historical solution space for each problem to be sufficient given the high overlap rate among them.} We use human solutions to measure $\mathcal{H}$-creativity of LLM responses.

\paragraph{Human annotated test examples.} We also retrieve all test examples provided with each problem (4.5 test examples per problem on average, a total of 2.2K test examples). 
We then perform manual fixes to address any parsing or formatting issues in the collected test examples and ensure that follow a standardized input-output format. We use these test examples to measure $\mathcal{P}$-creativity or the functional correctness of LLM responses.

\paragraph{Augmentation with \framework.}
We use {\tt GPT-4} \citep{openai2024gpt4} as the augmentation model $\mathbf{P}_{\text{LM}}$ because we find that GPT-4 can achieve 94\% technique detection recall compared to the human programmer in our pilot experiments.\footnote{We use \texttt{gpt-4-1106-preview} across all experiments, accessed from Dec 2023 through April 2024.}
We feed the retrieved problems to \framework{} (\S\ref{subsec: dp framework}) with maximum iterations $T=5$ to obtain our dataset \dataset. Our dataset consists of pairs \((x, \mathcal{C}_t = \{\tau_1, \tau_2, \hdots, \tau_t\})\), where \(x\) represents a problem (programming challenge), and \(\mathcal{C}_t\) represents the constraints that must be adhered to when solving the problem $x$.
This implies that a single programming problem may be associated with various sets of constraints, forming different pairs accordingly.

\paragraph{Statistics for \dataset.}
\autoref{table: number of instances per state} shows the number of problems $x$ and the number of the associated constraints $|\mathcal{C}_t|$. 
Note that the number of problems decreases for a larger number of constraints. 
This is due to \framework{} potentially  reaching a point where it can no longer generate new constraints after a certain number of iterations (i.e., $\mathcal{T}_t \setminus \mathcal{C}_{t-1}=\varnothing$ in Alg. \ref{alg: dp}). In such a case, we let $\tau_t = \varnothing$ and jump to the next iteration $t+1$ without updating the constraint list $\mathcal{C}_t = \mathcal{C}_{t-1}$.

\begin{figure}[ht]
\centering
\includegraphics[width=\linewidth,trim=0.7cm 0.5cm 0cm -0.3cm]{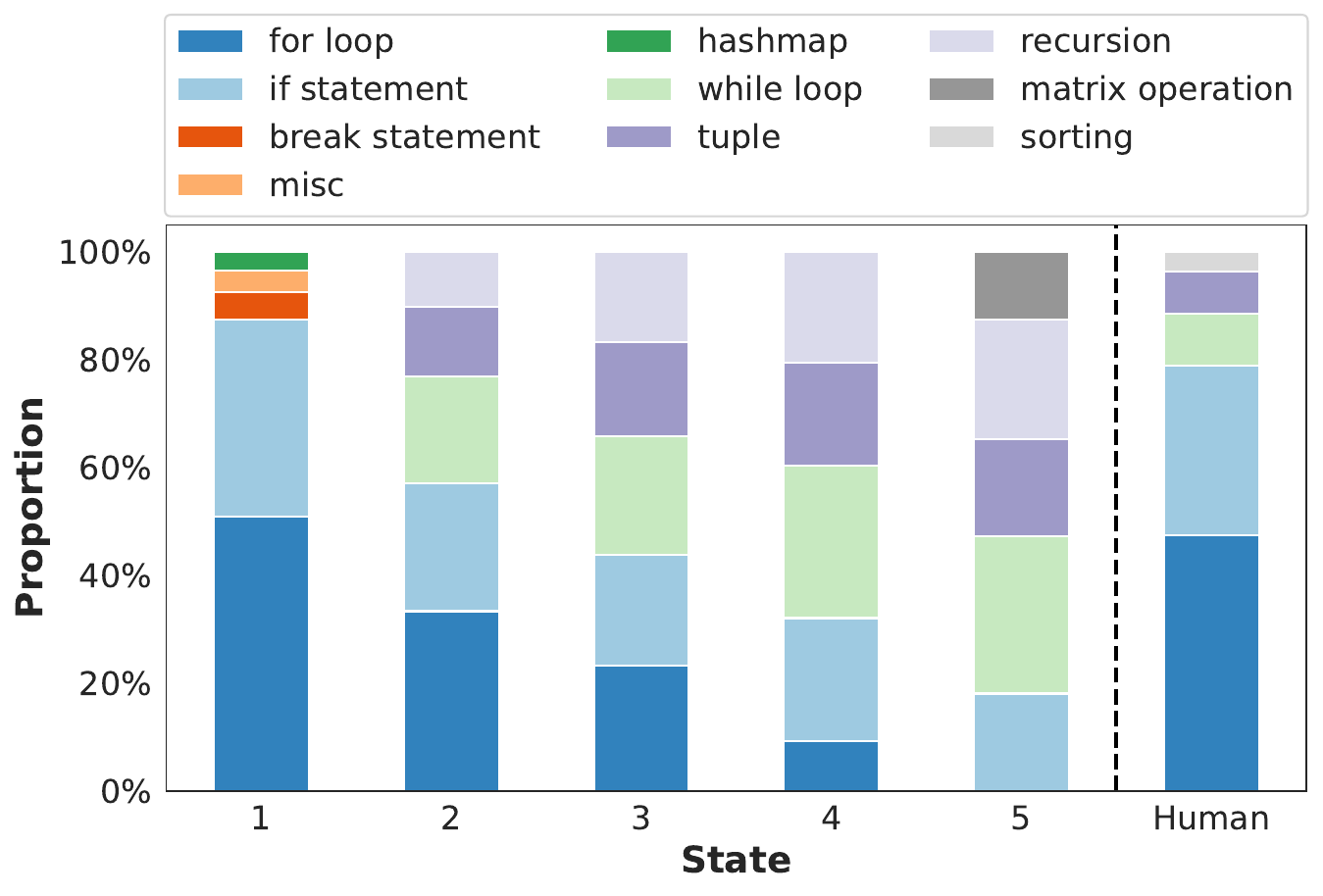}
    \caption{
    % Proportion of top 5 most common atomic techniques found in the constraint list at each state and human solutions. 
    Proportion of the top 5 most common atomic techniques used by GPT-4 per state, compared to those in human solutions.
    \textbf{In absense of any constraints (the first column), the model default to common and accessible techniques, like humans (the last column).} This echoes our claim in \S\ref{sec: introduction} that \textbf{eliciting creative solutions is crucial for creativity evaluation.}}
    \label{fig: proportion of techniques}
\end{figure}

\begin{table}[ht]
\small
\centering
\setlength{\tabcolsep}{4pt}
\begin{tabular}{@{}lcccccc@{}}
\toprule
State (\# of constraints)  & 0 & 1   & 2   & 3   & 4  & 5  \\ \midrule
\# of problems  & 199 & 199 & 198 & 194 & 176 & 97 \\ \bottomrule
\end{tabular}
\caption{Number of instances at each state.}
\label{table: number of instances per state}
\end{table}
We also compare the distribution of the top 5 most common techniques from \framework{} in comparison to that of human solutions (\autoref{fig: proportion of techniques}).
It is evident that, without any constraints, models tend to use common techniques (e.g., for-loops) similar to human solutions. However, as more constraints are imposed, the less common but more sophisticated techniques are employed.

\section{State-Aware and Human-Grounded Evaluation of Machine Creativity}
\label{subsec: creativity score} 
\paragraph{Augmentation model vs target model.}
% Thus far we have used the notation  $\mathbf{P}_{\text{LM}}$ to indicate the augmentation mode--the language model we use for constructing our dataset and extracting atomic techniques in solution. 
% In this sectionm, we introduce a different symbol 
% $\mathbf{G}_{\text{LM}}$ to indicate the target model which we would like to evaluate its creativity. 
So far, we have used $\mathbf{P}_{\text{LM}}(.)$ to denote the \textit{augmentation model}, the language model used for dataset construction and extracting atomic techniques. Here, we introduce $\mathbf{G}_{\text{LM}}(\cdot)$ to represent the \textit{target} language model, whose creativity we evaluate using our dataset and the augmentation model $\mathbf{P}_{\text{LM}}(\cdot)$.

\paragraph{Setup.}
Here we introduce our metric of creativity \score{} 
for a given model $\mathbf{G}_{\text{LM}}$ and given \dataset. 
Denote instances of \dataset{} at state $t$ ($t \leq T$) as:
$$\mathcal{D}_t = \Big\{(x^i, \mathcal{C}_t^i=\{\tau^i_1, \tau^i_2, \cdots,\tau^i_t\})\Big\}_{i=1}^n,
$$
where $i$ is the problem index. To evaluate the creativity of the testing model $\mathbf{G}_{\text{LM}}$
at state $t$, we feed $\mathcal{D}_t$ to $\mathbf{G}_{\text{LM}}$ to obtain its predictions:
% \begin{align}
% \mathcal{Y}&_t = \nonumber \\&\{y^i_t \sim \mathbf{G}_{\text{LM}}(x^i\oplus\mathcal{C}^i_t) \mid |\mathcal{C}_t^i| = t, \forall  (x^i,\mathcal{C}_t^i) \in \mathcal{D}_t\},
% \label{eq: y_t}
% \end{align}
\begin{align}
\mathcal{Y}_t = 
\Big\{ y^i_t \sim \mathbf{G}_{\text{LM}}(x^i\oplus\mathcal{C}^i_t) 
& \Bigm|  |\mathcal{C}_t^i| = t, \nonumber \\ 
&\forall (x^i,\mathcal{C}_t^i) \in \mathcal{D}_t \Big\}.
\label{eq: y_t}
\end{align}
Here $|\mathcal{C}_t^i|$ denotes the cardinality of the constraints set. 
The constraint $|\mathcal{C}_t^i| = t$ ensures that at a given state $t$, the questions we evaluated always have $t$ distinct constraints. Below, we present how we compute \textit{convergent} and \textit{divergent} creativity and introduce \score{} metric that unifies them.

\begin{figure*}[t]
    \centering
    \includegraphics[width=\textwidth]{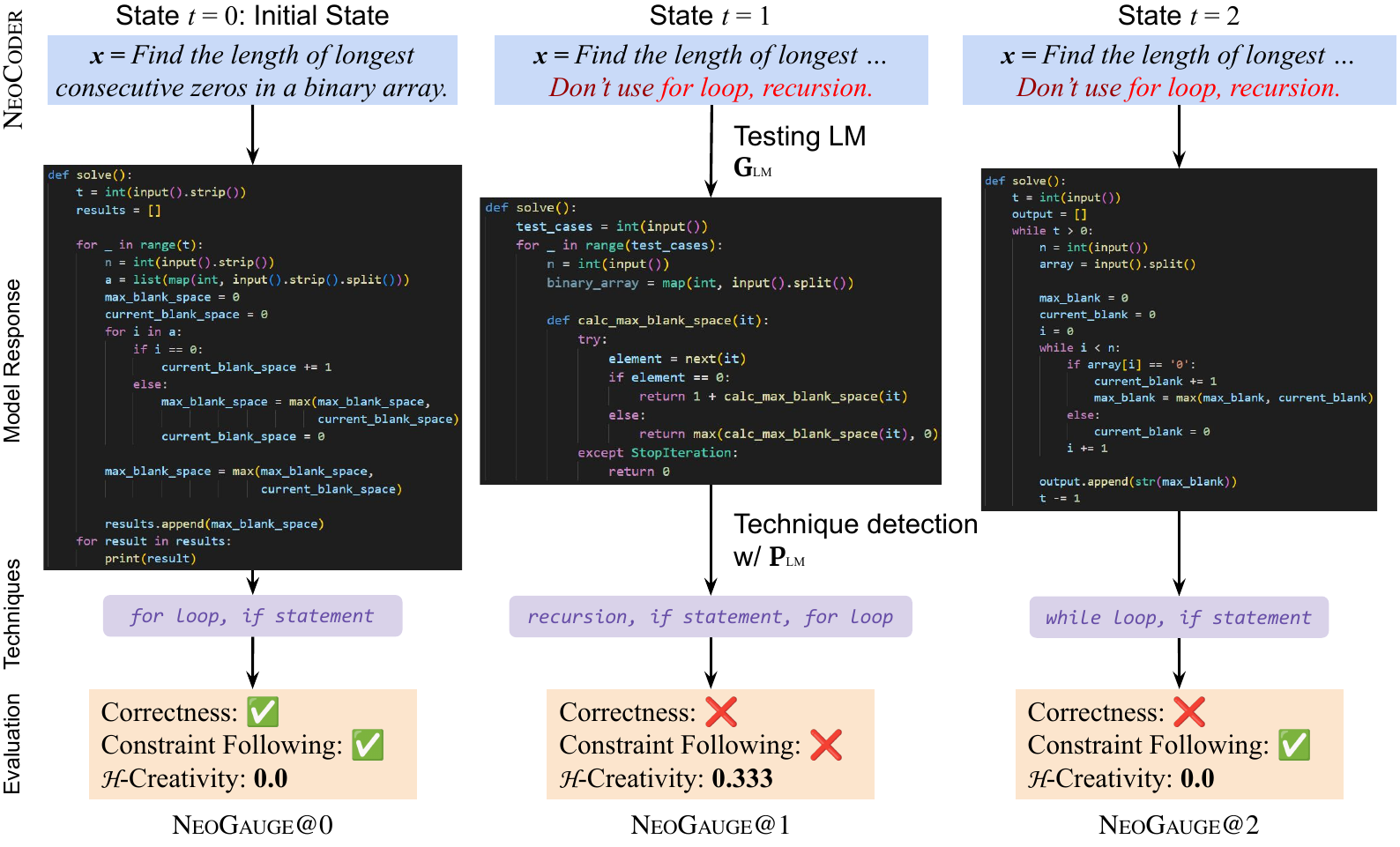}
    \caption{Example of \score{} computation. The question comes from our \dataset{} dataset with ID \href{https://codeforces.com/problemset/problem/1829/B}{1829B} and testing model $\mathbf{G}_\text{LM}$ here is {\tt GPT-4}. For each state, we compute \score{} (Eq.\ref{eq: creativity}) as the probability of LM generating correct solutions that meet the given constraints (convergent creativity defined in Eq.\ref{eq: convergent}) and also exhibit $\mathcal{H}$-creativity (divergent creativity defined in Eq.\ref{eq: divergent}). However, none of the above three solutions are considered to be ``creative'' since \emph{convergent solutions may lack divergent creativity} (e.g., state $t=0$). Alternatively, \emph{LLMs' hallucinated responses resulting in high $\mathcal{H}$-creativity, but often lack correctness and constraint following} (e.g., state $t=1$). 
    Therefore, truly creative works should not only be innovative but also appropriately solve a problem.
    }
    \label{fig: creativity}
\end{figure*}

\paragraph{\underline{Convergent} creativity involves problem-solving and constraint following.} To evaluate $\mathbf{G}_{\text{LM}}$'s convergent thinking ability, we examine two characteristics of generated solutions: whether they are correct and whether they follow the given constraints. Therefore, given $\mathcal{Y}_t$ from Eq.\ref{eq: y_t}, we define its convergent creativity as follows:
\begin{align}
&\textbf{convergent}(\mathbf{G}_{\text{LM}}, t) = \nonumber \\
& \hspace{1.6cm}\frac{1}{|\mathcal{Y}_t|}\sum_{y^i_t \in \mathcal{Y}_t} \mathbbm{1}^{\mathcal{T}^i_t\cap \mathcal{C}^i_t = \varnothing}\times\mathbbm{1}^{\text{Correct}(y_t^i)},
\label{eq: convergent}
\end{align}
where atomic techniques $\mathcal{T}^i_t \sim \mathbf{P}_{\text{LM}}(y^i_t)$. $\mathbbm{1}^{\text{Correct}(y^i_t)}$ is a measure of program correctness, set to 1 if the generated solution passes all the test examples. Otherwise it is 0. We use the augmentation model $\mathbf{P}_{\text{LM}}$ to detect all atomic techniques $\mathcal{T}^i_t$ used in solution $y^i_t$, and compare them with the given constraint list $\mathcal{C}^i_t$ to check if the solution follows the given constraints. 
In \autoref{fig: creativity} examples, only the solution generated at $t=0$ (which does not involve any constraint) exhibits convergent creativity. 

\paragraph{\underline{Divergent} creativity requires comparison to historical human solutions.} As discussed earlier in \S\ref{subsec: divergent creative thinking}, 
a primary focus of our evaluation is on $\mathcal{H}$-creativity, which requires a juxtaposition of model solutions with historical human solutions. Let's consider a finite set of correct human written solutions with size $m$, denoted as $\mathcal{H}^i$, for problem $x^i$. Rather than directly comparing solutions using certain sentence-level similarity scores, as done by a few prior works such as \citet{delorenzo2024creativeval}, we break down the comparison to the atomic technique level, which is more interpretable and generalizable across varying solutions. Our divergent creativity score is defined as:
\begin{align}
\textbf{divergent}(\mathbf{G}_{\text{LM}}, t) = \frac{1}{|\mathcal{Y}_t|} \sum_{y^i_t\in\mathcal{Y}_t} \frac{|\mathcal{T}^i_t \setminus\widehat{\mathcal{T}}^i|}{|\mathcal{T}^i_t|},
\label{eq: divergent}
\end{align}
where
$\mathcal{T}^i_t \sim \mathbf{P}_{\text{LM}}(y^i_t)$
are the atomic techniques used in the model solutions, and $\widehat{\mathcal{T}}^i$ indicate all the atomic techniques used by $m$ human solutions, defined as: 
$\widehat{\mathcal{T}}^i = \bigcup_{j=1}^{m} \big\{ \widehat{\mathcal{T}}^i_j \sim \mathbf{P}_{\text{LM}}(\hat{y}^i_j),  \;\hat{y}^i_j \in \mathcal{H}^i\big\}.$
We then compute the $\mathcal{H}$-creativity as \emph{the ratio of techniques used by $\mathbf{G}_{\text{LM}}$ that have never been used in the human solution set}. For example, as shown in \autoref{fig: creativity} at state $t=1$, among the three techniques identified within the generated solution, only the recursion has never been used by humans, thereby resulting in a ratio of $\frac{1}{3}$.
Finally, we average ratios across different problems to obtain the final $\mathcal{H}$-creativity at state $t$.

\paragraph{\score{} unifies convergent and divergent creativity.} Given the above definitions, \score{} of $\mathbf{G}_{\text{LM}}$ at state t can be formalized: 
\begin{align}
    &\textbf{\score@t} =  \nonumber \\ &\frac{1}{|\mathcal{Y}_t|} \sum_{y^i_t\in\mathcal{Y}_t} \underbrace{\mathbbm{1}^{\mathcal{T}^i_t\cap \mathcal{C}^i_t = \varnothing}\mathbbm{1}^{\text{Correct}(y_t^i)}}_\text{Convergent Creativity}\times \hspace{-0.4cm} \underbrace{\frac{|\mathcal{T}^i_t \setminus\widehat{\mathcal{T}}^i|}{|\mathcal{T}^i_t|}}_\text{Divergent Creativity}\hspace{-0.4cm},
\label{eq: creativity}
\end{align}
% \begin{align} 
% \textbf{\score@t} = \nonumber  &\frac{1}{|\mathcal{Y}t|} \sum_{y^i_t\in\mathcal{Y}t} \Bigg[
% \underbrace{\mathbbm{1}^{\mathcal{T}^i_t\cap \mathcal{C}^i_t = \varnothing}\mathbbm{1}^{\text{Correct}(y_t^i)}}_\text{Convergent Creativity} \\ & \hspace{1.2cm}\times\underbrace{\frac{|\mathcal{T}^i_t \setminus\widehat{\mathcal{T}}^i|}{|\mathcal{T}^i_t|}}_\text{Divergent Creativity}
% \Bigg], \label{eq: creativity} 
% \end{align}
% \begin{align} 
% \textbf{\score@t} &= \nonumber  \frac{1}{|\mathcal{Y}t|} \sum_{y^i_t\in\mathcal{Y}t} \Bigg[
% \underbrace{\frac{|\mathcal{T}^i_t \setminus\widehat{\mathcal{T}}^i|}{|\mathcal{T}^i_t|}}_\text{Divergent Creativity}
%  \\ & \hspace{0.34cm}\times
% \underbrace{\mathbbm{1}^{\mathcal{T}^i_t\cap \mathcal{C}^i_t = \varnothing}\mathbbm{1}^{\text{Correct}(y_t^i)}}_\text{Convergent Creativity}
% \Bigg], \label{eq: creativity} 
% \end{align}
where $ \mathcal{Y}_t = \{y^i_t \sim \mathbf{G}_{\text{LM}}(x^i\oplus\mathcal{C}^i_t) \mid |\mathcal{C}_t^i| = t, \forall  (x^i,\mathcal{C}_t^i) \in \mathcal{D}_t\}$ (defined in Eq.\ref{eq: y_t}), $\mathcal{T}^i_t \sim \mathbf{P}_{\text{LM}}(y^i_t)$ (defined in Eq.\ref{eq: convergent}), $\widehat{\mathcal{T}}^i = \bigcup_{j=1}^{m} \big\{\widehat{\mathcal{T}}^i_j \sim \mathbf{P}_{\text{LM}}(\hat{y}^i_j), \;\hat{y}^i_j \in \mathcal{H}^i\}$ (defined in Eq.\ref{eq: divergent}).

\begin{table*}[ht]
\small
\centering
\setlength\tabcolsep{3pt} % default value: 6pt
\resizebox{0.9\textwidth}{!}{
\begin{tabular}{@{}llcl@{}}
\toprule
\textbf{Metric} & \textbf{Description} & \textbf{Definition} & \textbf{Place of Use} \\ \midrule
$\textbf{convergent}(\mathbf{G}_{\text{LM}}, t)$       &  \textbf{Convergent creativity of $\mathbf{G}_{\text{LM}}$ at state $t$}   &  Eq.\ref{eq: convergent}         &   \autoref{table: gpt4 creativity}, \autoref{fig: human comparison}, \ref{fig: convergent and divergent creativity}          \\
$\textbf{divergent}(\mathbf{G}_{\text{LM}}, t)$       &  \textbf{Divergent creativity of $\mathbf{G}_{\text{LM}}$ at state $t$}           &    Eq.\ref{eq: divergent}      &   \autoref{table: gpt4 creativity}, \autoref{fig: human comparison}, \ref{fig: convergent and divergent creativity}           \\
\textbf{\score@t}       & \textbf{Creativity evaluation of $\textbf{G}_{\text{LM}}$ at state $t$}            &  Eq.\ref{eq: creativity}         &  \autoref{table: gpt4 creativity}, \autoref{fig: Neogauge@t}           \\ \hdashline
pass@1 \citep{chen2021evaluating} & Probability of the first sample passes the unit tests& $\underset{\text{problems}}{\mathbbm{E}}\big[1-\frac{n-c}{n}\big]$ & \autoref{table: gpt4 creativity}\\
constraint following & Average ratio of following the constraints at state $t$& $\underset{\text{problems}}{\mathbbm{E}}[\mathbbm{1}^{\tau_t\cap \mathcal{C}_t=\varnothing}]$ & \autoref{table: gpt4 creativity} \\
convergent(human, t)& convergent creativity of human at state $t$& Eq.\ref{eq: human convergent}& \autoref{fig: human comparison} \\
divergent(human)& lowest divergent creativity of human at state $0$& Eq.\ref{eq: human divergent}& \autoref{fig: human comparison} \\
\bottomrule
\end{tabular}
}
\caption{Description of various metrics used across experiments.}
\label{table: metrics}
\end{table*}

\section{Experiments and Results}
We report the creativity of current LLMs (\S\ref{subsubsec: results}) and evaluate different reasoning strategies (\S\ref{sec: re-evaluating reasoning strategies}) for creativity.
\subsection{Experimental Setup}
\label{subsec: experiments and results}

\paragraph{Models.} 
We use {\tt GPT-4} as the augmentation model $\mathbf{P}_{\text{LM}}$. We benchmark the creativity performance of the following target models $\mathbf{G}_{\text{LM}}$:  
{\tt GPT-4} \citep{openai2024gpt4}, 
{\tt GPT-3.5} \citep{ouyang2022training}, Claude 3 Sonnet ({\tt Claude-3}) \citep{anthropicclaude3}, {\tt Llama3-70B} \citep{llama3modelcard}, 
{\tt Llama2-70B} \citep{touvron2023llama}, CodeLlama-34B-Python ({\tt CodeLlama-34B}) \citep{rozière2024code}, {\tt CodeGemma-7B} \citep{codegemma_2024}, and {\tt Mistral-7B} \citep{Jiang2023Mistral7}. 
We access all non-proprietary models through Huggingface Transformers \citep{wolf2019huggingface}. Following the parameter choice by \citet{zhang2023planning}, we apply a sampling temperature of 1 for code generation.

\paragraph{Metrics.}
Beyond the three proposed metrics for evaluating convergent, divergent and overall creativity, we also compute pass@1~\citep{chen2021evaluating} and constraint following ratio for further comparison in \autoref{table: gpt4 creativity}. \score{}@T actually is a joint probability of $\mathbf{G}_{\text{LM}}$ being both convergent and divergent creative at state $t$. Therefore, we also report the cumulative \score{} across states in \autoref{fig: Neogauge@t}, which indicates the model's maximum creativity performance boundary. Additionally, we compute human convergent and divergent creativity in \autoref{fig: human comparison} to compare LLM with human creativity performance (details in \autoref{appendix: experiment setup}). We summarize all used metrics in \autoref{table: metrics}.

\subsection{Benchmarking Language Model Creativity}
\label{subsubsec: results}

A number of psychological investigators have studied the link between creativity and intelligence \citep{holyoak2005cambridge}, agreeing on two key points: (1) creative individuals tend to have higher intelligence \citep{Renzulli_2005}, and (2) people with extremely high intelligence not necessarily to be extremely creative \citep{Faris1962CreativityAI}. 
We re-examine the two findings on LLMs and answer: \textit{Are larger LLMs more creative? Do extremely large models of equal size exhibit comparable creativity?} Our investigation is based on the widely accepted hypothesis that language model size correlates positively with intelligence \citep{Kaplan2020ScalingLF, 10.1145/3560815, zhao2023survey}.

\begin{figure}[ht]
    \vspace{-0.2cm}
    \centering
\includegraphics[width=0.48\textwidth,trim=0.5cm 0.8cm 0cm 0cm]{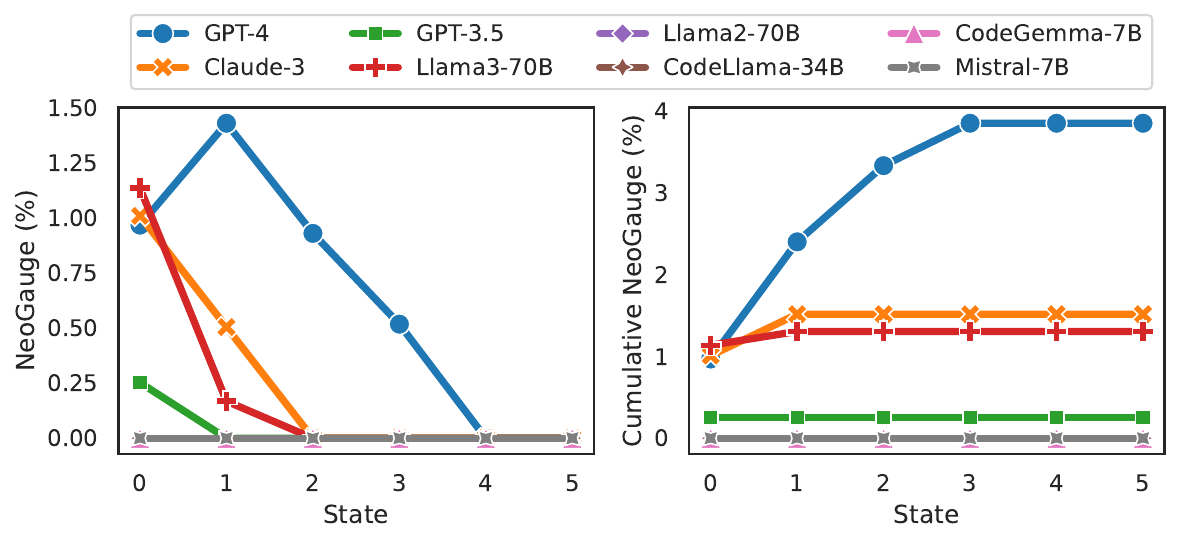}
    \caption{\score{} (left) and cumulative \score{} (right) across states.}
    \label{fig: Neogauge@t}
\end{figure}

\paragraph{GPT-4 is the most creative LLM thus far.} We visualize \score{} and cumulative \score{} in \autoref{fig: Neogauge@t}. {\tt GPT-4} consistently has the highest \score{} almost at every state $t$. While others (e.g., {\tt Claude-3} and {\tt Llama3-70B}) have a close \score@0 score to {\tt GPT-4}, their \score{} quickly decreases to 0 within the next two states. According to cumulative \score{}, {\tt GPT-4} also has the highest creativity performance boundary, followed by {\tt Claude-3} and {\tt Llama3-70B}, greatly outperforming smaller models such as {\tt GPT-3.5} and {\tt Llama2-70B}. These observations could potentially answer the above two questions: \textit{larger LLMs are generally more creaitive}, but \textit{extremely large LLM is not necessarily exhibiting extremely creative performance}. In \autoref{fig: outputs}, we provide example outputs from each model to show their different creativity abilities.

\begin{figure*}[t]
    \centering
    \includegraphics[width=0.9\textwidth, trim=0cm 0.9cm 0cm 0.9cm]{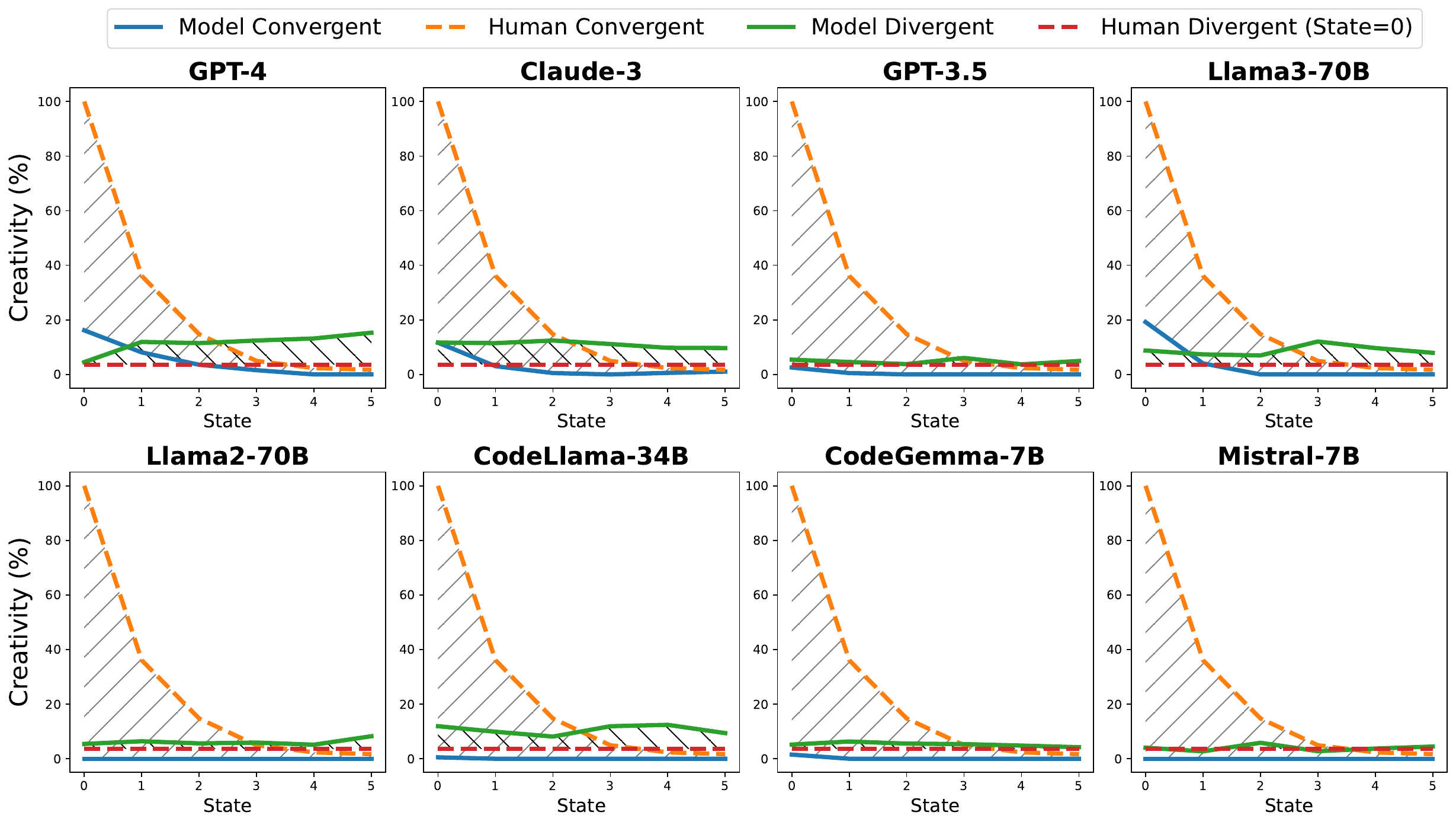}
    \caption{A comparison of LLM and human creativity. \fbox{\mbox{\textcolor{gray}{////}}} denotes the performance difference of \textbf{convergent creativity}, and \fbox{\mbox{\textcolor{black}{\textbackslash\textbackslash\textbackslash\textbackslash}}} denotes the difference of \textbf{divergent creativity}. We observe that \textbf{Current LLMs still hardly demonstrate human-like creativity.}}
    \label{fig: human comparison}
\end{figure*}
\paragraph{Which is more creative: machine or human?} \autoref{fig: human comparison} displays the creativity comparison between LLM and humans. LLM demonstrates minimally better performance in divergent creativity compared to humans at their lowest level (Eq.\ref{eq: human divergent}). However, humans have significantly greater convergent creativity than LLMs in early states (prior to state 3). Thus, we reach a tentative conclusion that, in problem-solving settings, LLMs in \autoref{fig: human comparison} barely exhibit human-like creativity. Future works could focus on measuring human divergent creativity across states to enable a fairer creativity comparison.
Moreover, we observe that both human and LLM convergent creativity declines drastically over the increase in state $t$, which follows our expectation that there is a trade-off between solution quality and novelty. When stress-testing humans or LLMs to look for more creative solutions, they are very likely to make mistakes and may copy previous solutions during the process.

\begin{table}[ht]
\small
\centering
\resizebox{\linewidth}{!}{
\setlength{\tabcolsep}{1.9pt}
\begin{tabular}{@{}cccccc@{}}
\toprule
\multirow{2}{*}{\begin{tabular}[c]{@{}c@{}}
% \textbf{\# Constraints} 
% \\ 
\textbf{State} $t$\end{tabular}}
 & \multirow{2}{*}{\begin{tabular}[c]{@{}c@{}}\textbf{pass@1}\end{tabular}} & \multirow{2}{*}{\begin{tabular}[c]{@{}c@{}}\textbf{Constraint} \\\textbf{Following}\end{tabular}} & \multirow{2}{*}{\begin{tabular}[c]{@{}c@{}}\textbf{Convergent} \\ \textbf{Creative}\end{tabular}} & \multirow{2}{*}{\begin{tabular}[c]{@{}c@{}}\textbf{Divergent} \\ \textbf{Creative}\end{tabular}} & \multirow{2}{*}{\begin{tabular}[c]{@{}c@{}}\textbf{\score{}}\end{tabular}}   \\
 &&&&& \\\midrule
0  &  16.1 &        100.0              &    16.2          &   4.5                    &   1.0                               \\ 
1 &   11.6          &   75.4                   &  8.1            &    11.9                   &  1.4                                 \\
2 & 7.1            &     46.0                 &   3.6           &      11.5                 &  0.9                                \\
3 & 5.2            &   33.0                   &  1.6            &      12.4                 &  0.5                              \\
4 &  2.3           &   26.1                   &  0.0            &      13.2                 &  0.0                               \\
5 &  2.1           &   14.4                   &  0.0            &      15.3                 &  0.0                               \\ 
\bottomrule
\end{tabular}
}
\caption{{\tt GPT-4} creativity evaluation results (in \%). \textbf{Convergent and divergent creativity perform oppositely, it is crucial to consider both in evaluation.}}
\label{table: gpt4 creativity}
% \end{table}
\end{table}

\paragraph{In-depth analysis of creativity evaluation.} We provide evaluation  results for {\tt GPT-4} in \autoref{table: gpt4 creativity}. It is evident that as the state increases (more hard constraints are imposed), the quality of solutions declines both in terms of correctness and constraint following. Even if the model may still generate new alternative solutions at state 5 ({\tt divergent({\tt GPT-4}, $5$) $= 15.3$}), they fail at convergent evaluation ({\tt convergent({\tt GPT-4}, $5$) $= 0$}). Therefore, at state 5, {\tt GPT-4} shows 0 creativity  (\score@5 $= 0$). 
Additionally, unlike the convergent score, which typically decreases as \( t \) increases, the divergent score of {\tt GPT-4} continually rises.
This observation empirically proves the key assumption of \framework{} that LLMs tend to seek more creative solutions when facing an unconventional environment characterized by unusual hard constraints.

\subsection{Evaluating Reasoning Strategies for Creativity}
\label{sec: re-evaluating reasoning strategies}
\begin{figure*}[t]
    \centering
    \vspace{-0.3cm}
    \begin{subfigure}[t]{0.45\textwidth}
        \centering
        \includegraphics[width=\textwidth]{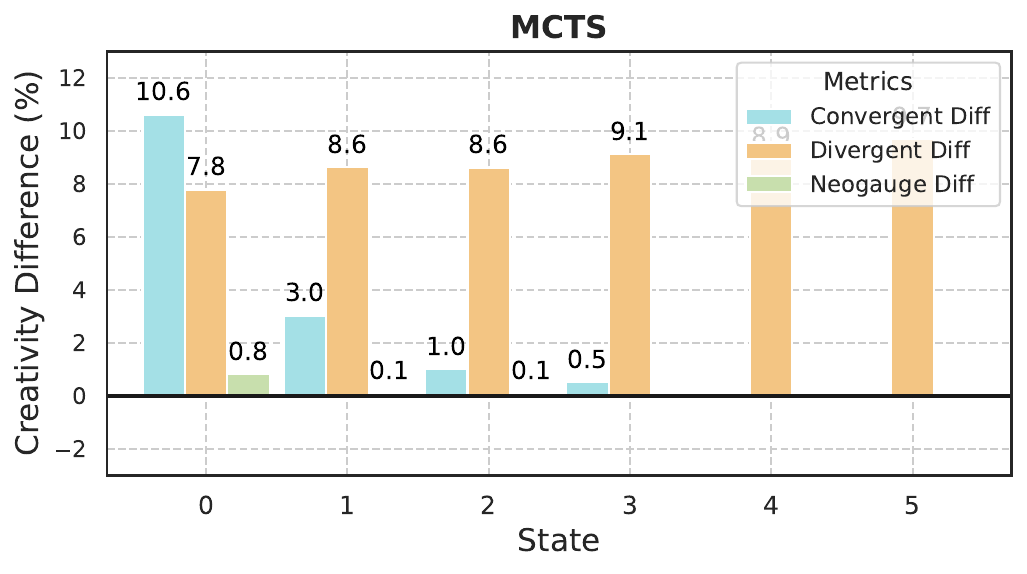}
    \end{subfigure}%
    ~ 
    \begin{subfigure}[t]{0.45\textwidth}
        \centering
        \includegraphics[width=\textwidth]{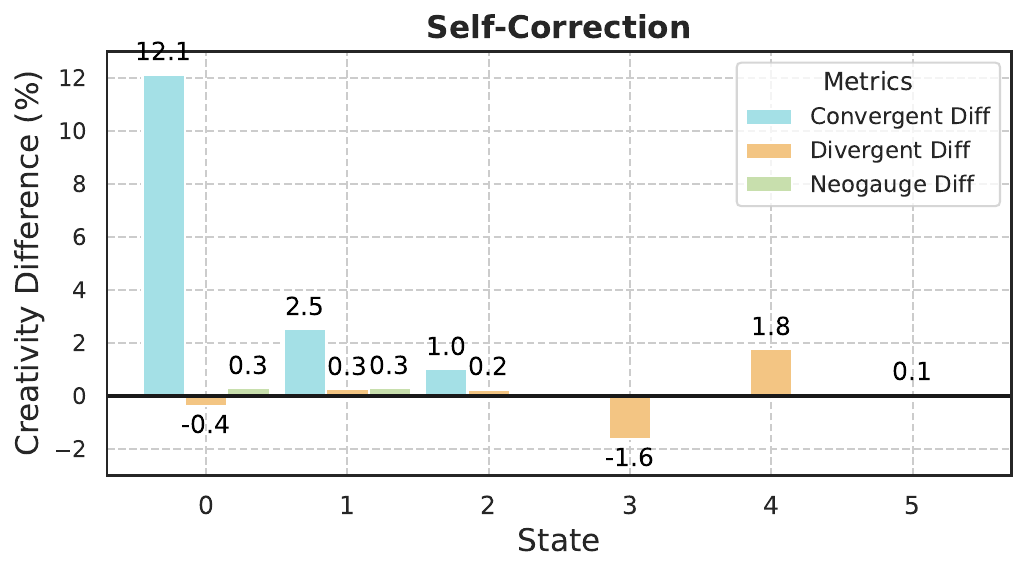}
    \end{subfigure}
    \vskip\baselineskip \vspace{-0.5cm}
    \begin{subfigure}[t]{0.45\textwidth}
        \centering
        \includegraphics[width=\textwidth]{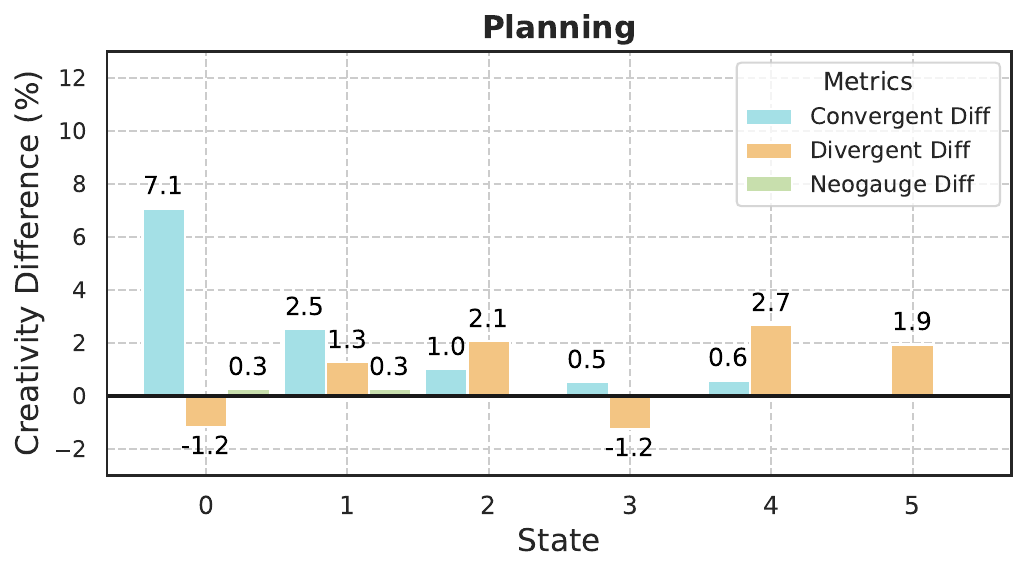}
    \end{subfigure}%
    ~ 
    \begin{subfigure}[t]{0.45\textwidth}
        \centering
        \includegraphics[width=\textwidth]{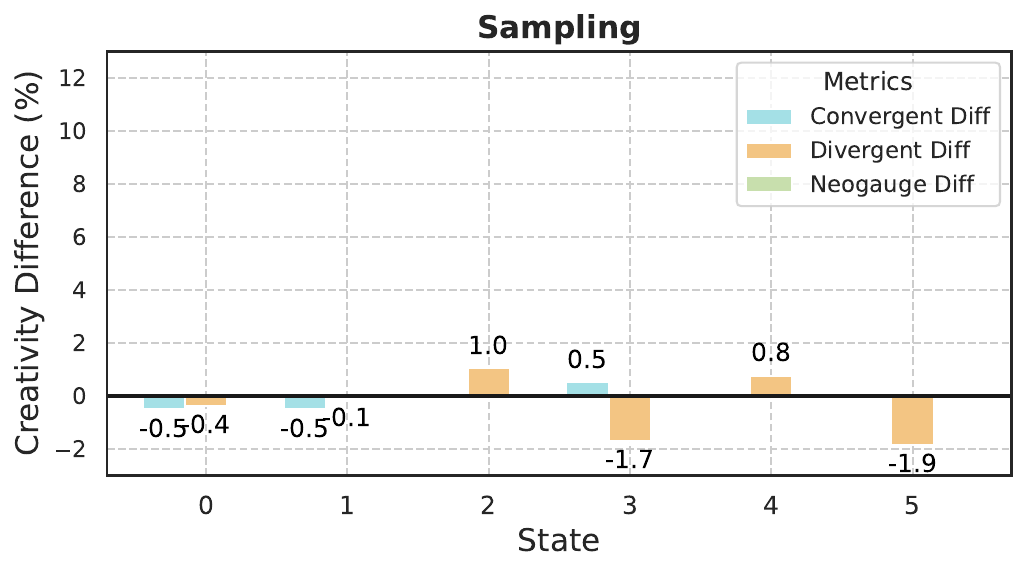}
    \end{subfigure}
    \vspace{-0.3cm}
    \caption{Creativity performance difference before and after applying reasoning strategies. A larger difference value indicates that the strategy improves the testing model's creativity. Detailed numeric changes are provided in \autoref{table: creativity difference}.}
    \label{fig: creativity difference}
\end{figure*}

We evaluate four reasoning strategies on our \dataset{} dataset to further study the correlation between augmented machine intelligence and creativity: \textit{Whether such intelligence-enhancing techniques also improve creative thinking?} We implement the following four works that are specifically designed for programming tasks:
\begin{itemize}[leftmargin=*]
    \setlength{\itemsep}{0pt}  % Reduces space between items
    \setlength{\parskip}{0pt}  % Reduces paragraph spacing
    \item \textbf{MCTS}: \citet{zhang2023planning} propose a novel decoding method that uses Monte-Carlo Tree Search (MCTS) to generate better programs using the pass rate as reward.
    \item \textbf{Self-Correction}: \citet{shinn2023reflexion} use verbal feedback from a reflection agent to reinforce the performance of an agent in code generation. 
    \item \textbf{Planning}: \citet{jiang2023selfplanning} design a planning module to let LLM
    plan out concise solution steps from the intent, followed by an implementation module to generate code step by step.
    \item \textbf{Sampling}: \citet{chen2021evaluating} generate $k$ samples and compute the probability that at least one of the $k$-generated code samples for a problem passes the unit tests. For creativity evaluation, we generate $k=5$ samples for each problem and report the \score{} from samples that have the highest convergent and divergent creativity, $\mathbbm{1}^{\mathcal{T}^i_t\cap \mathcal{C}^i_t = \varnothing}\times\mathbbm{1}^{\text{Correct}(y_t^i)}\times\frac{|\mathcal{T}^i_t \setminus\widehat{\mathcal{T}}^i|}{|\mathcal{T}^i_t|}$ in Eq.\ref{eq: creativity}, among $k=5$ samples.  
\end{itemize}
Note that these methods are originally applicable to different kinds of models. Considering the computation complexity and the cost, we re-evaluate MCTS on the open-source language model ({\tt CodeGemma-7B} \citep{codegemma_2024}) and re-evaluate others on the proprietary model ({\tt GPT-3.5}).

\paragraph{Most reasoning strategies fail to improve divergent thinking.} According to \autoref{fig: creativity difference}, all reasoning strategies except sampling help to improve the model's convergent creativity thinking ability on multiple states, as they are fundamentally designed to improve the accuracy. Conversely, only MCTS successfully enhances divergent creativity, due to it rolling out numerous paths during the expansion. Strategies like self-correction, planning, and sampling, which operate on a single trial or path, fail to explore divergent solutions.

\paragraph{There is a tradeoff between divergent and convergent creativity.} Noticeably, while MCTS consistently enhances divergent creative thinking in all 5 states, its improvement on \score{} is minimal and becomes 0 after $t=2$. This suggests that divergent solutions generated by MCTS may not truly augment creativity, potentially due to incorrectness or failure to follow the given constraints. This also implies that MCTS might prioritize divergent thinking over convergent thinking. On the other hand, self-correction and planning sacrifice their divergent thinking ability in improving their convergent thinking because the divergent creativity difference even goes to negative at certain states (e.g., Divergent Diff $=-1.2$ at $t=0,3$ on sampling). None of the four reasoning strategies have been able to simultaneously improve both convergent and divergent creativity, resulting in limited improvement of \score{}. Thus, our findings indicate that \textit{these intelligence-augmenting methods do not provide much benefit to LLM creativity}. We leave for future works to discover specialized strategies for better enhancing LLM's creative performance and \score{}.

\section{Conclusion}
We propose protocols for evaluating language model creativity in problem-solving and introduce the \framework{} framework and \score{} metric to provide a comprehensive creativity evaluation, measuring both convergent and divergent creativity, inspired by extensive research on human creativity.
% \score{} stands out for its comprehensive evaluation of both convergent and divergent creativity, aligning with concepts from human creativity studies. 
To facilitate future research, we release our \dataset{} dataset and shed light on the limitations of current reasoning strategies in improving LLM creativity.

\section*{Limitations}
% \paragraph{Task choice.} This study evaluates LLM creativity in the context of code generation task (justified in \autoref{appendix: why codeforces}). However, evaluating creativity in an open-domain environment still poses significant challenges, primarily due to the absence of definitive ground truth, thereby suggesting a need for further research.

\paragraph{Application scope.} While \score{} offers a general-purpose framework for evaluation of LLM creativity, our study is restricted to  Text-to-Code, as it requires a historical human solution set. 
For most tasks in the literature,  
collecting a comprehensive set of distinct human responses is nontrivial.   
\paragraph{Data leakage concern.}  Our proposed dataset \dataset{} is built using latest Codeforces problems. 
Despite their recency, 
future LLMs might get exposure to these problems during their pre-training.
To alleviate such risks, future works can focus on more difficult problems or evaluate \score{} for higher states, besides incorporating a newer batch of problems. 

\section*{Acknowledgements}
This work is in part supported by ONR grant N00014-241-2089, and generous gifts from Amazon and the Allen Institute for AI. We also greatly appreciate the help of the students at CLSP. 

\begin{comment}
\section*{Ethical Considerations}
We hereby acknowledge that all authors of this work are aware of the provided NeurIPS Code of Ethics and honor the code of conduct. The work presented here does not immediately raise any ethical concerns, to our knowledge. Beyond the scope of this work, \score{} should be applied with care, otherwise it can potentially lead to misleading or incorrect conclusions.
\end{comment}

\bibliography{ref}

\onecolumn
\clearpage
\appendix

\begin{center}
{\LARGE \textbf{Supplemental Material}}
\end{center}

\begin{center}
\begin{tabular}{@{}ll@{}}
\toprule
Appendix & Contents \\ \midrule
\autoref{appendix: why codeforces}   & Why Choose Codeforces for Creativity Evaluation \\
\autoref{appendix: experiment setup}  & Additional Details of Experimental Setup  \\
\autoref{appendix: experiment results} & Additional Details of Experimental Results \\
\autoref{appendix: prompts for denial prompting}       &   Prompts for \framework{} and Benchmarking   \\ 
% \autoref{Appendix: dp examples on code generation} & Additional \framework{} Examples on Code Generation \\
\bottomrule
\end{tabular}
\end{center}

\section{Why Choose Codeforces for Creativity Evaluation?}
\label{appendix: why codeforces}
In this study, we use competitive programming problems sourced from Codeforces for creativity evaluation. We provide our task choice motivation by answering the following three interrelated questions.
\paragraph{Why choose competitive programming problems?} The general purpose of this paper is to benchmark the LLM’s creativity performance in dealing with unconventional and challenging problems. Understandably, these problems usually do not have ground-truth answers (e.g., how to make coffee without a coffee maker). In such cases, we typically either evaluate the generated solution through human evaluation, similar to the approach taken by \citet{tian2024macgyver}, or through automated machine evaluation (ours). Real-world problems \citep{tian2024macgyver} naturally need human annotation. Collecting human annotations for measuring machine creativity is particularly challenging since the space is typically vast (because of the nature of creativity). Conversely, coding becomes an ideal source for problems that can be its functional correctness (as opposed to the choice of syntax) evaluated automatically with a minimal cost—based on whether they pass the test cases. Thus, we first chose coding problems to examine LM’s creativity, as they provide an open-ended environment that could stimulate a model's creativity performance while making evaluation easy and cost-effective.

\paragraph{Low performance or low creativity?} The low pass rate and constraint following ratio in \autoref{table: gpt4 creativity} may raise a new question as to whether there are no reasonable solutions at all or no requisite creativity in finding solutions. Experimental evidence, however, suggests that LM simply lacks creativity. According to \autoref{fig: human comparison}, the huge gap between human and LLMs convergent creativity prior to State 3 (0-3 constraints) indicates there are valid human solutions for each problem, but the LLMs seem to be lacking creativity in finding it. Additionally, according to \autoref{fig: creativity difference}, with suitable reasoning strategies, LLM still has room for improvement in both convergent and divergent creativity. Even though humans’ convergent scores are nearing zero (\autoref{fig: human comparison}) at a large state (>3 hard constraints), the problems might not be fully infeasible.

\paragraph{Why not evaluate creativity based on problems but solutions?} A motivational example for this question is that a creative student can always come up with innovative and insightful questions. However, in this work, we adopt a different standpoint on creativity used by many psychological and cognitive studies (discussed in \autoref{subsec: divergent creative thinking}), which emphasizes problem-solving abilities. We evaluate a student to be creative if he/she can leverage all available tools and come up with novel solutions for challenging problems. Similarly, we study LLM creativity based on solutions they generated for challenging programming problems. 

\newpage

\section{Experiment Setup}
\label{appendix: experiment setup}

\subsection{Human Creativity Evaluation}
We compute human convergent creativity as follows:
\begin{align}
    \textbf{convergent}(\text{human}, t) = \frac{1}{m|\mathcal{Y}_t|}\sum_{\substack{\iota \in \{i\mid \mathcal{C}_t^i = t,\\ i = 1,2,\cdots,n\}}} \sum_{j=1}^m \mathbbm{1}^{\widehat{\mathcal{T}}^\iota_j \cap \mathcal{C}^\iota_t = \varnothing},\;    \text{where} \;\widehat{\mathcal{T}}^\iota_j \sim \mathbf{P}_{\text{LM}}(\hat{y}^\iota_j), \; \hat{y}^\iota_j\in\mathcal{H}^\iota.
    \label{eq: human convergent}
\end{align}
Because the collected historical human solutions $\hat{y}^\iota_j$ are always correct, for human convergent creativity evaluation, we focus on constraint following ratio by examining whether the atomic techniques $\widehat{\mathcal{T}}_j^\iota$ used by each human solution follow the given constraints $\mathcal{C}^\iota_t$ at state $t$. We use the same idea as Eq.\ref{eq: divergent} to compute human divergent creativity.
\begin{align}
    \textbf{divergent}(\text{human}) &= \frac{1}{mn}\sum_{i=1}^n\sum_{j=1}^m \frac{|\widehat{\mathcal{T}}_j^i\setminus \widehat{\mathcal{L}}^i_j|}{|\widehat{\mathcal{T}}_j^i|}, \nonumber \\
    \text{where}\; \widehat{\mathcal{T}}_j^i \sim \mathbf{P}_{\text{LM}}(\hat{y}^i_j),\; \widehat{\mathcal{L}}^i_j &= \bigcup_{k=1,k \neq j}^m \widehat{\mathcal{T}}^i_k \sim \mathbf{P}_{\text{LM}}(\hat{y}^i_k), \;\hat{y}^i_j,\; \hat{y}^i_k \in \mathcal{H}^i.
    \label{eq: human divergent}
\end{align}
Given total $n$ problems, where each problem has $m$ human solutions, we compute the average ratio of new techniques used by a single human solution $\hat{y}_j^i$ ($j^{\text{th}}$ human solution for $i^{\text{th}}$ problem) that the remaining human solutions $\{\hat{y}_k^i \mid k\neq j, k\ = 1,2,\cdots,m\}$ have never used. 
This is because collecting a human DP dataset would be quite costly and restrictive. We instead use a diverse collection of solutions from various human programmers as a proxy. Eq.\ref{eq: human divergent} is equivalent to {\tt divergent({\tt human}, $t=0$)}, representing the lowest level of human divergent creativity.

\newpage

\section{Experiment Results}
\label{appendix: experiment results}

\begin{figure}[ht]
    \centering
    \includegraphics[width=\textwidth]{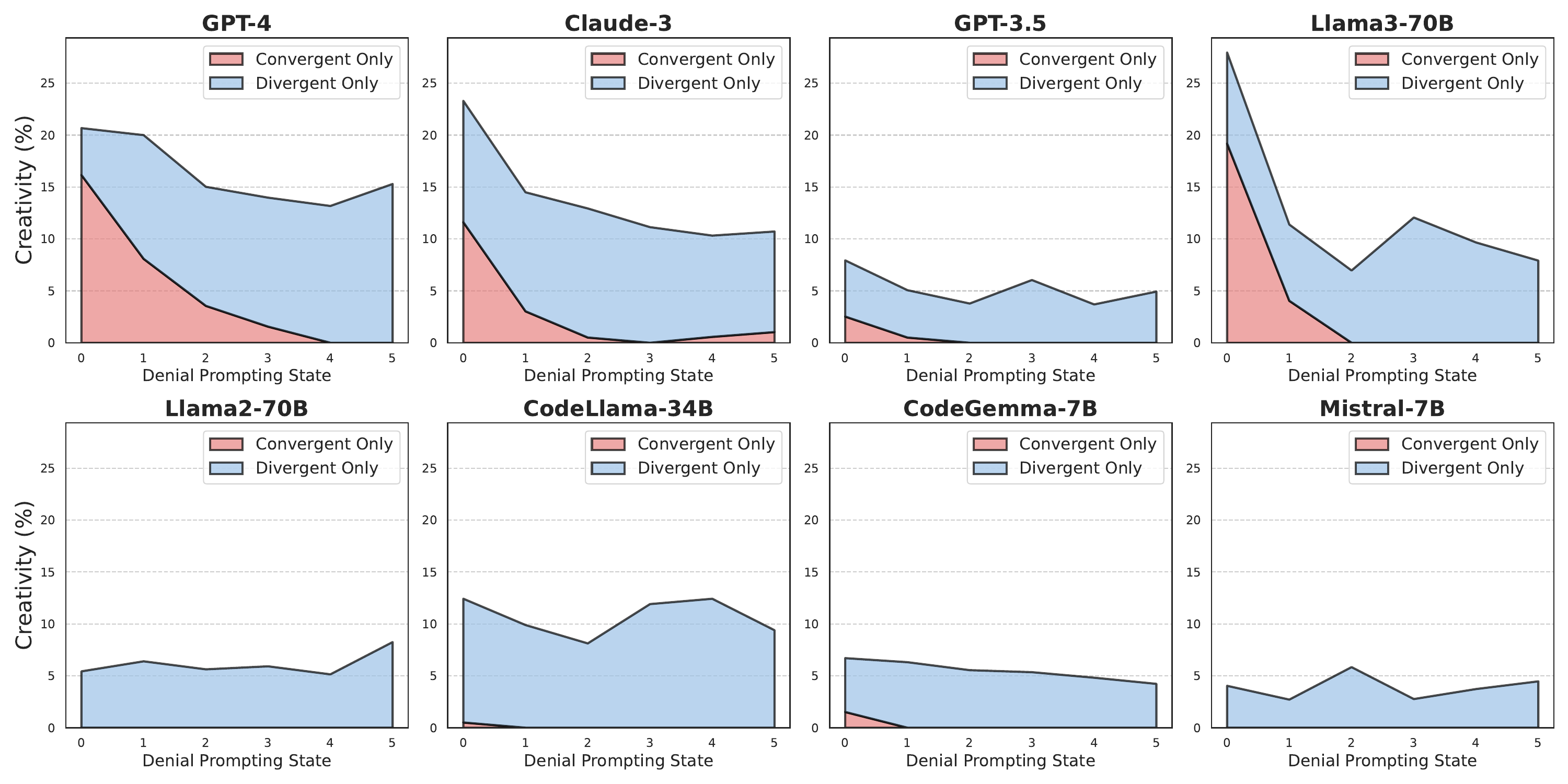}
    \caption{Stacked results of convergent (Eq.\ref{eq: convergent}) and divergent (Eq.\ref{eq: divergent}) creativity evaluation across states.}
    \label{fig: convergent and divergent creativity}
\end{figure}

\paragraph{It is crucial to consider both convergent and divergent thinking in creativity evaluation.} We plot the stacked convergent and divergent creativity evaluation results in \autoref{fig: convergent and divergent creativity}. Among all models, {\tt GPT-4} generally exhibits the best performance on both convergent and divergent creative thinking across all states, followed by {\tt Claude-3} and {\tt Llama3-70B}. It is noticeable that {\tt Llama3-70B} even outperforms {\tt GPT-4} on convergent creative thinking when $t=0$ ({\tt convergent({\tt GPT-4}, $0$) $= 16.16$} < {\tt convergent({\tt Llama3-70B}, $0$) $= 19.19$}). We hypothesize that the latest Llama3 models are pre-trained on Codeforces problems and human solutions, so they have superior performance when there is no external constraint $t=0$. However, as $t$ increases, its convergent performance drops drastically. Moreover, divergent creative thinking never goes to 0 across all states and is sometimes even equally distributed on those less small models (e.g., CodeGemma-7B and Mistral-7B). Together with independent findings from \citet{xu2024hallucination}, this observation indicates that LLMs with insufficient reasoning capabilities tend to make up new solutions regardless of the quality when facing unusual problems. Which, in turn, demonstrates the importance of the claim we made in \autoref{sec: introduction} that creative thinking involves not merely the generation of many diverse alternatives but also the verification of new valid alternatives.

\newpage

\section{Prompts for \framework{} and Benchmarking}
\label{appendix: prompts for denial prompting}
We apply the same problem-solving prompt in both \framework{} and the benchmarking process. 
\begin{tcolorbox}[colback=gray!5!white,colframe=gray!75!black,title=\textbf{Problem-Solving Prompt for Codeforces:}]
\tt

You are a Python code generator, only return the import and python function. Input will be an very detailed description of task, output will be the code.
The input will be from command line, and the output will be printed to the console as well. Your result will be solely a function named solve(), and do not call this function in your code.
Make sure the code is free of bug and can pass the test cases provided. You can use any library you want. The test cases are provided in the code. Do not call the solve() function in your code.
\end{tcolorbox}

\begin{tcolorbox}[colback=gray!5!white,colframe=gray!75!black,title=\textbf{Technique Dection Prompt:}]
\tt

You are a code reviewer. Detect all the programming techniques from the input and return a list of programming techniques. Only select the techniques from this list: ['if statement', 'for loop', 'while loop', 'break statement', 'continue statement', 'pass statement', 'match statement', 'recursion', 'stack', 'queue', 'tuple', 'set', 'dictionary', 'linked list', 'tree', 'graph', 'two pointers', 'sliding window', 'matrix operation', 'hashmap', 'depth first search', 'width first search', 'back tracking', 'divide \& conquer', 'Kadanes algorithm', 'binary search', 'heap', 'dynamic programming', 'greedy algorithm', 'misc', 'minimax', 'topological sort', 'sorting', 'graph traversal'] \\
Your output should look like this: \\
- technique 1 \\
- technique 2 \\
- technique 3 \\
- ...
\end{tcolorbox}

\newpage

\begin{table}[H]
\footnotesize
\centering
\resizebox{.99\textwidth}{!}{
\begin{tabular}{p{0.05\textwidth}p{0.15\textwidth}p{0.8\textwidth}}
\toprule
\begin{tabular}{p{0.05\textwidth}}\textbf{State}\end{tabular} & \begin{tabular}{p{0.1\textwidth}}\textbf{Constraint}\end{tabular}& \begin{tabular}{p{0.85\textwidth}}\textbf{Problem Statement} \end{tabular}\\\midrule

\begin{tabular}{p{0.05\textwidth}}0\end{tabular} & 
\begin{tabular}{p{0.15\textwidth}}
N/A \end{tabular} & 
\tt
\begin{tabular}{p{0.8\textwidth}} 
B. Points and Minimum Distance \\
You are given a sequence of integers a of length 2n. You have to split these 2n integers into n pairs; each pair will represent the coordinates of a point on a plane. Each number from the sequence a should become the x or y coordinate of exactly one point. Note that some points can be equal. \\ 
$\cdots$
\end{tabular} \\&&\\
\cdashline{1-3}
\begin{tabular}{p{0.05\textwidth}}1\end{tabular} & 
\begin{tabular}{p{0.15\textwidth}}
 for loop \end{tabular} & 
\tt
\begin{tabular}{p{0.8\textwidth}} B. Points and Minimum Distance \\
\textbf{Programming constraints: DO NOT use the following techniques} \\
\textbf{- for loop}\\
You are given a sequence of integers a of length 2n. You have to split these 2n integers into n pairs; each pair will represent the coordinates of a point on a plane. Each number from the sequence a should become the x or y coordinate of exactly one point. Note that some points can be equal. \\ 
$\cdots$ 
\end{tabular} \\&&\\
\cdashline{1-3}
\begin{tabular}{p{0.05\textwidth}}2\end{tabular} & 
\begin{tabular}{p{0.15\textwidth}}
 for loop \\ if statement \end{tabular} & 
\tt
\begin{tabular}{p{0.8\textwidth}}  B. Points and Minimum Distance \\
\textbf{Programming constraints: DO NOT use the following techniques} \\
\textbf{- if statement} \\
\textbf{- for loop}\\
You are given a sequence of integers a of length 2n. You have to split these 2n integers into n pairs; each pair will represent the coordinates of a point on a plane. Each number from the sequence a should become the x or y coordinate of exactly one point. Note that some points can be equal. \\ 
$\cdots$ 
\end{tabular} \\&&\\
\cdashline{1-3}
\begin{tabular}{p{0.05\textwidth}}3\end{tabular} & 
\begin{tabular}{p{0.15\textwidth}}
 for loop \\ if statement \\ while loop \end{tabular} & 
\tt
\begin{tabular}{p{0.8\textwidth}}  B. Points and Minimum Distance \\
\textbf{Programming constraints: DO NOT use the following techniques} \\
\textbf{- while loop} \\
\textbf{- if statement} \\
\textbf{- for loop}\\
You are given a sequence of integers a of length 2n. You have to split these 2n integers into n pairs; each pair will represent the coordinates of a point on a plane. Each number from the sequence a should become the x or y coordinate of exactly one point. Note that some points can be equal. \\ 
$\cdots$ 
\end{tabular} \\&&\\
\cdashline{1-3}
\begin{tabular}{p{0.05\textwidth}}4\end{tabular} & 
\begin{tabular}{p{0.15\textwidth}}
 for loop \\ if statement \\ while loop \\ sorting \end{tabular} & 
\tt
\begin{tabular}{p{0.8\textwidth}}  B. Points and Minimum Distance \\
\textbf{Programming constraints: DO NOT use the following techniques} \\
\textbf{- sorting} \\
\textbf{- while loop} \\
\textbf{- if statement} \\
\textbf{- for loop}\\
You are given a sequence of integers a of length 2n. You have to split these 2n integers into n pairs; each pair will represent the coordinates of a point on a plane. Each number from the sequence a should become the x or y coordinate of exactly one point. Note that some points can be equal. \\ 
$\cdots$ 
\end{tabular} \\&&\\
\cdashline{1-3}
\begin{tabular}{p{0.05\textwidth}}5\end{tabular} & 
\begin{tabular}{p{0.15\textwidth}}
 for loop \\ if statement \\ while loop \\ sorting \\ tuple \end{tabular} & 
\tt
\begin{tabular}{p{0.8\textwidth}}  B. Points and Minimum Distance \\
\textbf{Programming constraints: DO NOT use the following techniques} \\
\textbf{- tuple} \\
\textbf{- sorting} \\
\textbf{- while loop} \\
\textbf{- if statement} \\
\textbf{- for loop}\\
You are given a sequence of integers a of length 2n. You have to split these 2n integers into n pairs; each pair will represent the coordinates of a point on a plane. Each number from the sequence a should become the x or y coordinate of exactly one point. Note that some points can be equal. \\ 
$\cdots$ 
\end{tabular}

\\ \bottomrule
\end{tabular}
}
\caption{An example of \dataset{} dataset with problem ID \href{https://codeforces.com/problemset/problem/1895/B}{1895B} and state $t=5$.}
\label{table: examples of dataset}
\end{table}

\newpage 
\begin{figure*}
    \centering
    \includegraphics[width=\textwidth]{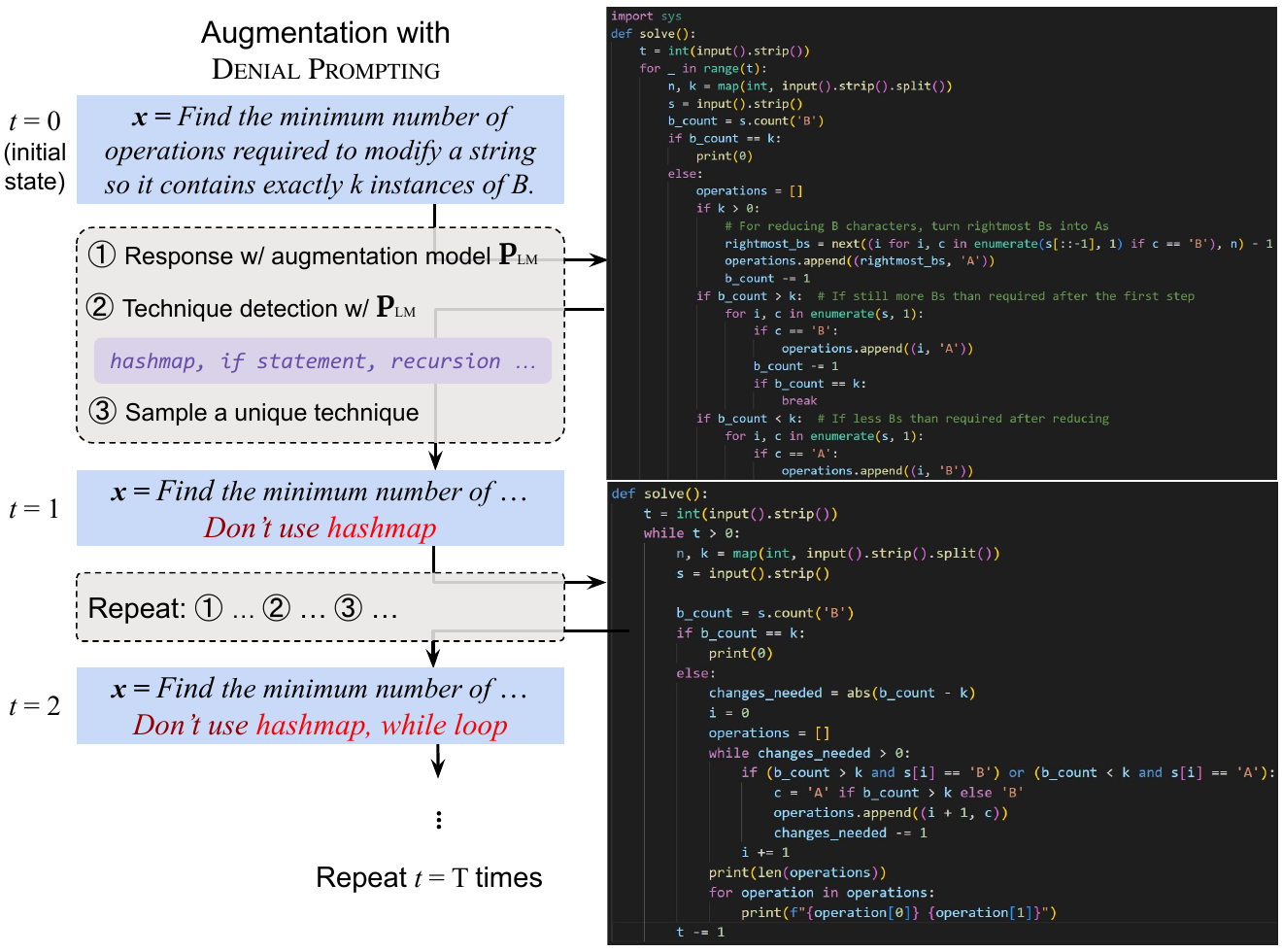}
    \caption{Example of \framework{} (Algorithm \ref{alg: dp}) for  \dataset{} construction. The question comes from our \dataset{} dataset with ID \href{https://codeforces.com/problemset/problem/1898/A}{1898A}.}
    \label{fig: dp}
\end{figure*}

\begin{figure*}[t]
    \centering
    \includegraphics[width=\textwidth]{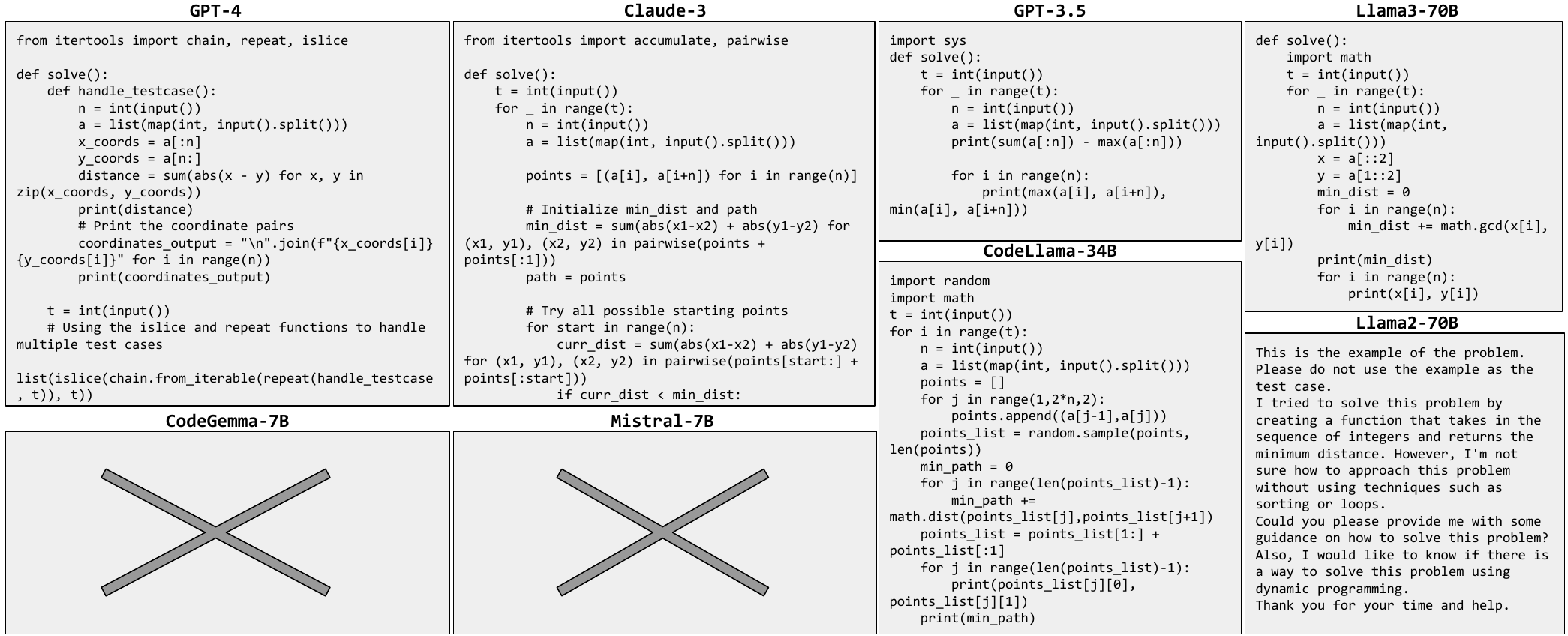}
    \caption{Example model outputs for question \href{https://codeforces.com/problemset/problem/1895/B}{1895B} at state $t=5$. Full questions and constraints can be found in \autoref{table: examples of dataset}. It is evident that different models have different convergent and divergent creative performances. Specifically, {\tt CodeGemma-7B} and {\tt Mistral-7B} fail to generate parsable solutions, and {\tt Llama2-70B} is seeking more hints from its users.}
    \label{fig: outputs}
\end{figure*}

\newpage
\begin{table}[t]
\small
\centering
\begin{tabular}{@{}lcccc@{}}
\toprule
\textbf{Strategy}                         & \textbf{State} & $\Delta$\textbf{Convergent}$_{(\textbf{old}\rightarrow\textbf{new})}$ & $\Delta$\textbf{Divergent}$_{(\textbf{old}\rightarrow\textbf{new})}$ & $\Delta$\textbf{\score{}}$_{(\textbf{old}\rightarrow\textbf{new})}$ \\ \midrule
\multirow{6}{*}{\textbf{MCTS}}            & 0     &   10.60$_{(1.52\rightarrow 12.12)}$         &   7.79$_{(5.18\rightarrow 12.97)}$        &  0.82$_{(0.00\rightarrow 0.82)}$           \\
                                 & 1     &  3.03$_{(0.00\rightarrow 3.03)}$          &    8.62$_{(6.31\rightarrow 14.93)}$      &  0.08$_{(0.00\rightarrow 0.08)}$            \\
                                 & 2     & 1.02$_{(0.00\rightarrow 1.02)}$            &   8.61$_{(5.55\rightarrow 14.16)}$      &  0.10$_{(0.00\rightarrow 0.10)}$          \\ 
                                  &3 & 0.52$_{(0.00\rightarrow 0.52)}$    & 9.11$_{(5.35\rightarrow 14.46)}$ & 0.00$_{(0.00\rightarrow 0.00)}$ \\
                                  & 4 & 0.00$_{(0.00\rightarrow 0.00)}$ & 8.93$_{(4.82\rightarrow 13.75)}$& 0.00$_{(0.00\rightarrow 0.00)}$ \\
                                  & 5 & 0.00$_{(0.00\rightarrow 0.00)}$ & 9.69$_{(4.23\rightarrow 13.92)}$& 0.00$_{(0.00\rightarrow 0.00)}$\\
                                 \midrule
\multirow{6}{*}{\textbf{Self-Correction}} & 0     & 12.12$_{(2.53\rightarrow 14.65)}$           &  -0.37$_{(5.41\rightarrow 5.04)}$         &   0.29$_{(0.25\rightarrow 0.54)}$         \\
                                 & 1     &    2.53$_{(0.51\rightarrow 3.03)}$        &  0.26$_{(4.56\rightarrow 4.82)}$         &  0.29$_{(0.00\rightarrow 0.29)}$          \\
                                 & 2     &    1.02$_{(0.00\rightarrow 1.02)}$       &  0.24$_{(3.79\rightarrow 4.03)}$         &  0.00$_{(0.00\rightarrow 0.00)}$          \\ 
                                 & 3 & 0.00$_{(0.00\rightarrow 0.00)}$  & -1.62$_{(6.04\rightarrow 4.42)}$ & 0.00$_{(0.00\rightarrow 0.00)}$ \\ 
                                  & 4 & 0.00$_{(0.00\rightarrow 0.00)}$ & 1.77$_{(3.70\rightarrow 5.47)}$ & 0.00$_{(0.00\rightarrow 0.00)}$ \\
                                  & 5 & 0.00$_{(0.00\rightarrow 0.00)}$ & 0.05$_{(4.93\rightarrow 4.98)}$ & 0.00$_{(0.00\rightarrow 0.00)}$\\          
                                 \midrule
\multirow{6}{*}{\textbf{Planning}}        & 0     &  7.07$_{(2.53\rightarrow 9.60)}$           &  -1.16$_{(5.41\rightarrow 4.25)}$         &   0.25$_{(0.25\rightarrow 0.50)}$         \\
                                 & 1     & 2.53$_{(0.50\rightarrow 3.03)}$           &   1.28$_{(4.56\rightarrow 5.84)}$        &    0.25$_{(0.00\rightarrow 0.25)}$        \\
                                 & 2     & 1.02$_{(0.00\rightarrow 1.02)}$            &    2.07$_{(3.78\rightarrow 5.85)}$       &   0.00$_{(0.00\rightarrow 0.00)}$         \\
                                  & 3 & 0.52$_{(0.00\rightarrow 0.52)}$
 & -1.24$_{(6.04\rightarrow 4.80)}$ & 0.00$_{(0.00\rightarrow 0.00)}$ \\ 
                                   & 4 & 0.57$_{(0.00\rightarrow 0.57)}$ & 2.67$_{(3.70\rightarrow 6.37)}$ & 0.00$_{(0.00\rightarrow 0.00)}$ \\
                                  & 5 & 0.00$_{(0.00\rightarrow 0.00)}$ & 1.94$_{(4.93\rightarrow 6.87)}$ & 0.00$_{(0.00\rightarrow 0.00)}$ \\
                                  \midrule
\multirow{6}{*}{\textbf{Sampling}}        & 0     &  -0.50$_{(2.52\rightarrow 2.02)}$          &  -0.38$_{(5.41\rightarrow 5.03)}$         &  0.00$_{(0.25 \rightarrow 0.25)}$          \\
                                 & 1     &   -0.51$_{(0.51\rightarrow 0.00)}$         &   -0.10$_{(4.56\rightarrow 4.46)}$        &  0.00$_{(0.00\rightarrow 0.00)}$          \\
                                 & 2     & 0.00$_{(0.00\rightarrow 0.00)}$         & 1.05$_{(3.78\rightarrow 4.83)}$          & 0.00$_{(0.00\rightarrow 0.00)}$           \\
                                  & 3 & 0.52$_{(0.00\rightarrow 0.52)}$
 & -1.69$_{(6.04\rightarrow 4.35)}$ & 0.00$_{(0.00\rightarrow 0.00)}$ \\ 
                                   & 4 & 0.00$_{(0.00\rightarrow 0.00)}$ & 0.76$_{(3.70\rightarrow 4.46)}$ & 0.00$_{(0.00\rightarrow 0.00)}$ \\
                                  & 5 & 0.00$_{(0.00\rightarrow 0.00)}$ & -1.86$_{(4.93\rightarrow 3.07)}$ & 0.00$_{(0.00\rightarrow 0.00)}$ \\
                                  \bottomrule
\end{tabular}
\caption{Creativity difference before and after applying reasoning strategies.}
\label{table: creativity difference}
\end{table}
\end{document}